\documentclass[runningheads]{llncs}
\usepackage{graphicx}

\usepackage{tikz}
\usepackage{comment}
\usepackage{amsmath,amssymb} 
\usepackage{color}
\usepackage{indentfirst} 
\usepackage{float} 
\usepackage[accsupp]{axessibility}  

\begin{document}
\pagestyle{headings}
\mainmatter
\def\ECCVSubNumber{1394}  

\title{Point Primitive Transformer for Long-Term 4D Point Cloud Video Understanding}

\titlerunning{Point Primitive Transformer}
%
\author{Hao Wen\inst{1}$^{*}$ \and
Yunze Liu\inst{1}$^{*}$ \and
Jingwei Huang\inst{2}\and
Bo Duan \inst{2} \and
Li Yi\inst{1,3}}

\authorrunning{Wen. et al.}
%
\institute{Tsinghua University \\
\email{wenh19@mails.tsinghua.edu.cn, liuyzchina@gmail.com} \and
Huawei Technologies 
\email{\{huangjingwei6,duanbo5\}@huawei.com} \and
Shanghai Qi Zhi Institute \email{ericyi0124@gmail.com}}
\maketitle

\begin{abstract}
This paper proposes a 4D backbone for long-term point cloud video understanding. A typical way to capture spatial-temporal context is using 4Dconv or transformer without hierarchy. However, those methods are neither effective nor efficient enough due to camera motion, scene changes, sampling patterns, and complexity of 4D data. To address those issues, we leverage the primitive plane as mid-level representation to capture the long-term spatial-temporal context in 4D point cloud videos, and propose a novel hierarchical backbone named Point Primitive Transformer(PPTr), which is mainly composed of intra-primitive point transformers and primitive transformers. Extensive experiments show that PPTr outperforms the previous state of the arts on different tasks.
\keywords{Transformer; Primitive; Long-term Point Cloud Video}
\end{abstract}
\footnotetext{$^{*}$Equal contribution. Author ordering determined by coin flip.}

\section{Introduction}
Point cloud videos are ubiquitous in robots and AR systems that act as a window into our dynamically changing 3D world. Being able to record movements in the physical space, point cloud sequences play a key role in comprehending environmental changes and supporting interactions with the world, which can be hardly described by 2D images or static 3D point clouds. Therefore, an intelligent agent must process such a form of data precisely to better model the real world, adapt to environmental changes, and interact with them.

Despite its importance, processing point cloud sequences is a quite challenging task for machines that are largely determined by two aspects: effectiveness and efficiency. Effectiveness refers to the ability to capture long-term spatial-temporal structures. Due to camera motion, scene changes, occlusion changes, and sampling patterns, points between different frames are unstructured and inconsistent, making it difficult to effectively integrate different frames into the underlying spatio-temporal structure. Efficiency refers to how to efficiently process long point cloud videos with limited computing resources. The complexity and dimension of 4D data can easily cause memory and computation explosions. Both challenges grow dramatically as the length of the video increases.

One typical way to tackle the dynamics of point clouds videos is treating the point cloud video as a 4D volume~\cite{minkowski}, which applies 4D convolution directly after voxelization. It is computationally prohibitive when processing large scenes and long videos. Compared with transformer-based 4D backbones, pure convolution is less effective at capturing long-term spatio-temporal context. However, the existing transformer-based 4D backbone(P4Transformer~\cite{p4transformer}) also fails to solve the above challenges. The entire point cloud video still needs to be loaded into memory during the training process, which severely limits the length of the point cloud video (for example, a 24GB graphics card can only handle a synthia4D~\cite{synthia} point cloud video of 3 frames). Additionally, even though flat transformers may be able to capture long-term context theoretically, they are difficult to optimize as point numbers increase and usually do not provide much gain in dense prediction tasks, such as 4D semantic segmentation.

Based on the challenges described above, we have several key observations. First, considering the large variety of points, distance point cloud frames should not be extracted at the point level, as this is neither efficient nor effective. Second, a middle-level abstraction representing the underlying geometry spatially and temporally can be better suited for context modeling, which will not only alleviate the need to process raw points for better efficiency but also allow for easier association across frames for a more effective spatial-temporal structure. After revisiting the geometry processing literature, we choose primitive plane as a mid-level representation, which describes the underlying planar structures in a scene and tends to be much more stable across frames.


\begin{figure}[t]
    \centering
    \includegraphics[width=\linewidth]{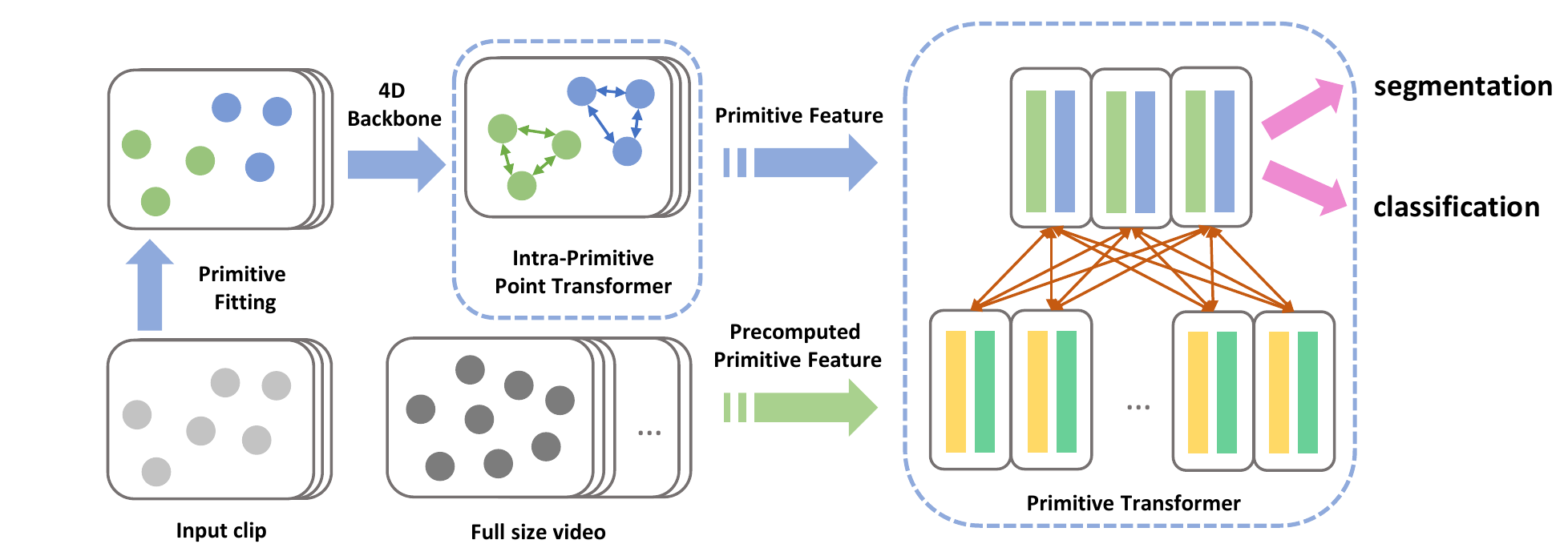}
    \caption{Architecture of Primitive Point Transformer. On the lower level, PPTr extracts short-term spatial-temporal features through an intra-primitive point transformer for a short video clip around the frame of interest. On the upper level, PPTr extracts long-term spatial-temporal features through a primitive transformer.}
    \label{fig:teaser}
\end{figure}

In this paper, we leverage primitive planes to develop an efficient and effective 4D backbone named Point Primitive Transformer(PPTr). As primitive planes induce a natural scene-primitive-point hierarchy in space, we also design PPTr as a hierarchical transformer operating on two different levels as shown in Figure~\ref{fig:teaser}. On the lower level, PPTr extracts short-term spatial-temporal features through an intra-primitive point transformer for a short video clip around the frame of interest. Primitive planes are used to restrict the spatial support of attention maps in a point-level transformer. Such geometry-aware locality inductive bias is not only beneficial for the optimization of the transformer but also very effective for extracting descriptive and temporally stable geometric features. On the upper level, PPTr extracts long-term spatial-temporal features through a primitive transformer. We allow very efficient consideration of a long sequence by fitting primitives and computing the primitive features in a pre-processing stage. Through the primitive transformer, we could better associate primitives from different frames and effectively integrate long-term context to the frame of interest. 

We evaluate our Point Primitive Transformer(PPTr) on several tasks, such as 3D action recognition on MSR-Action~\cite{msr} and 4D semantic segmentation on Synthia4D~\cite{synthia} and HOI4D~\cite{liu2022hoi4d}. we demonstrate significant improvements over previous method($+1.33\%$ mIoU on synthia4D, $+6.28\%$ mIoU on HOI4D and $+1.39\%$ accuracy on MSR-Action). 

The contributions of this paper are fourfold:
\begin{itemize}
\item First, we leverage the primitive plane to capture the long-term spatial-temporal context in 4D point cloud videos and propose a novel backbone named Point Primitive Transformer(PPTr). 
\item Second, we propose an intra-primitive point transformer for extracting spatially descriptive and temporally stable \textbf{short-term} geometric features. 
\item Third, we propose a primitive transformer to capture \textbf{long-term} spatial-temporal features efficiently. 
\item Fourth, extensive experiments on three datasets show that the proposed Point Primitive Transformer is more effective and efficient than previous state-of-the-art 4D backbones.
\end{itemize}

\section{Related Work}
\noindent{\bf Deep learning on Point Cloud Video Processing.} 
Different from grid-based RGB video, point cloud video exhibits irregularities and lacks order along the spatial dimension where points emerge inconsistently across time. One approach to deal with that is voxilization. For instance, ~\cite{minkowski} extends temporal dimension to 3D sparse convolution~\cite{spconv} to extract spatial temporal features on 4D occupancy grids. 3DV~\cite{3dv} proposes a 3D motion representation to encode 3D motion information via temporal rank pooling~\cite{rankpool}. Another approach is to perform directly on point sets. MeteorNet~\cite{meteornet} adopts PointNet++~\cite{pointnet2} to aggregate information from neighbors, while point-track is needed to merge points. PSTNet~\cite{pstnet} firstly decomposes spatial and temporal information and proposes a point-based convolution in a hierarchical manner. Following~\cite{pstnet}~\cite{meteornet}, P4Transformer~\cite{p4transformer} proposes 4D Convolution that performs spatial-temporal convolution and captures dynamics of points by self-attention. While like most point-based approaches, they prolong input clip by simply feeding raw points into network, which suffers from limited memory and fails to benefit from long-range temporal dependencies. Based on this, we propose Point Primitive Transformer(PPTr) which enjoys all three properties: point-convolution based, long-term supported and point-track avoided.

\noindent{\bf Primitive Fitting.} 
Primitive fitting is a long-standing problem of grouping points into specific geometric shapes such as plane, cuboid, cylinder and so on. Such process approximates and abstracts 3D shapes from low-level digitized point data to a succinct high-level parameterized representation. Two mainstream solutions of primitive fitting in geometry community are RANSAC~\cite{ransac}~\cite{efficientransac} and region grow~\cite{regiongrow1}~\cite{regiongrow2}. Recently, neural networks have been developed by several works~\cite{3dprnn}~\cite{spfn}~\cite{hpnet}~\cite{primitivenet}~\cite{volumetricprimitives} to segment primitives. Because primitives extremely simplifies point data while keeps a relatively precise description of 3D geometry, they are widely applied to downstream tasks like instance segmentation~\cite{primitivenet}, reconstruction~\cite{architecturalmodeling}and animation~\cite{animated}. For example, ~\cite{shapetemplates} utilizes primitive shapes that are rich in underlying structures to reconstruct scanned object and transfer the structural information onto new objects. To directly deal with large-scale scenes, ~\cite{SPG} distils organization of point cloud by partitioning heavy points into light shapes, showing the power of such compact yet rich representation. We inject primitives into our network, intending to spatially provide geometric-aware enhancement on local primitive region and temporally leverage long-range information in a memory efficient way.\\
{\bf Transformer Network.} 
Transformer is a powerful deep neural network based on self-attention mechanism~\cite{attention} and is particularly suitable for modelling long-range dependencies~\cite{nlp1}~\cite{nlp2}~\cite{nlp3}. It was firstly proposed in~\cite{attention} for machine translation task and further extended to vision community~\cite{vivit}~\cite{nonlocal3}~\cite{vit}~\cite{vit2}~\cite{nonlocal1}~\cite{nonlocal2}~\cite{nonlocal4} . Very recently, Swin Transformer~\cite{swin} proposes a hierarchical design for vision modeling at various scales and yields impressive results . Similar to CNNs~\cite{cnn1}~\cite{resnet}, Swin transformer builds hierarchical feature maps by merging image patches when layers go deeper, and strikes a balance between efficiency and effectiveness by limiting self-attention to local windows while also supporting cross-window connection. In 4D point cloud understanding, prior leading work~\cite{p4transformer} performs self-attention globally and fails to leverage long-term dependencies effectively. As such, we design a hierarchical Primitive Point Transformer(PPTr) to alleviate ineffectiveness of global-wise attention and introduce intra-primitive point transformer and primitive transformer that perform self-attention at point level and primitive level respectively. Intensive experiments have shown that our network outperforms the state-of-the-art methods for both 4D semantic segmentation and 4D action recognition.

\section{Pilot study: How does P4Transformer perform on long-term point cloud videos?}
4D point cloud video understanding has obtained much attention recently and researchers are actively seeking for backbones to capture descriptive spatial-temporal features. Among them, P4Transformer~\cite{p4transformer} is the leading one achieving state-of-the-art performance on common tasks including 4D semantic segmentation and 4D action recognition. Briefly speaking, instead of tracking points, P4Transformer uses a point 4D convolution to encode the spatio-temporal local structures in a point cloud video, and utilize the transformer to capture the global appearance and motion information across the entire video.
To motivate the necessity of a new backbone, we conduct a pilot study to understand the constraints of P4Transformer for long-term point cloud video understanding. 
\begin{figure}[t]
    \centering
    \includegraphics[width=\linewidth]{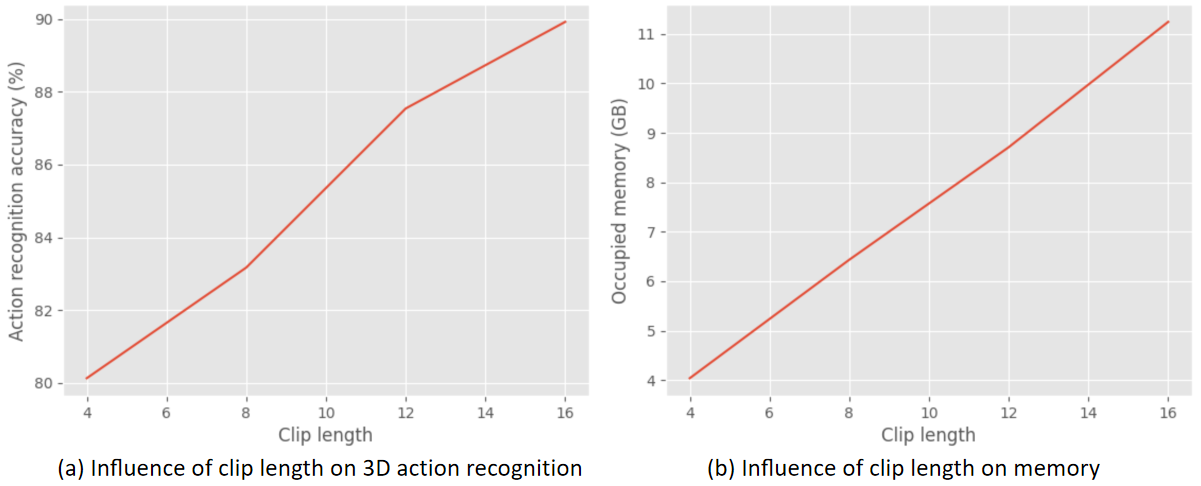}
    \vspace{-10pt}
     \caption{\textbf{(a)} The performance gain(MSR-Action3D~\cite{msr}) with the increase of temporal range. \textbf{(b)} The occupied memory with the increase of temporal range. We take the 2080Ti(11GB) GPU as an example. When the GPU memory cap is reached, the maximum number of frames that can be used is 15, which can only achieve $89\%$ accuracy. }
    \label{fig:pilot}
    \vspace{-10pt}
\end{figure}
\begin{itemize}
    \item We first experiment with the action recognition task on MSR-Action3D~\cite{msr} dataset. We gradually increase the clip length until our GPU memory cap is reached and examine how well P4transformer performs.
    The results are shown in Figure~\ref{fig:pilot}. 
    \item We further conduct 4D semantic segmentation experiments on the synthia 4D dataset~\cite{synthia}, to verify the effect of Transformer. Specifically, We removed the Transformer in P4Transformer and compared it with the full version.
\end{itemize}

We can draw mainly two conclusions from the above experiments. First, as shown in Figure~\ref{fig:pilot}, P4Transformer achieves better performance as the clip length increases but is soon restricted by the huge memory cost, and it is hard to apply P4Transformer to very long clips. When the GPU memory cap is reached, the performance still keeps its trend of going up, indicating the huge potential of exploring longer-term videos. 
Second, in synthia4D~\cite{synthia} semantic segmentation task, we find that P4transformer without Transformer can achieve mIoU of $80.3\%$, which only drops $2.86\%$ compared with original P4Transformer. This result indicates global spatial-temporal context captured by P4Transformer becomes less useful in 4D dense prediction tasks. This is quite counter-intuitive as the first conclusion indicates the benefit of modeling long-term information. We conjecture that using a flat transformer as in P4Transformer is not effective for long-term spatial-temporal context due to optimization issues.


We re-examine the design principles of 4D backbones for long-term videos and we would like to emphasize two important properties: efficiency (both speed-wise and memory-wise) and effectiveness. By efficiency, we mean the backbone should be able to effectively model long-term context to understand 4D visual data in a more integrated way. P4Transformer is not efficient since it needs to load a whole point cloud sequence into the memory for per-point feature learning. This could easily explode the memory as the sequence becomes longer or input scenes become larger-scale. Similar drawbacks also apply to most other 4D backbones in the literature. P4Transformer is also not effective enough for aggregating long-term context due to the usage of ball-like region features. P4Transformer samples equal-sized ball regions in each frame to compute feature tokens and applies transformer to a sequence of frames. The geometric meaning of such randomly sampled balls could hugely vary in dynamic scenes. This makes it hard to build long-term associations, which is important for long-term context.


\section{Method}

\begin{figure}[htbp]
    \centering
    \includegraphics[width=\linewidth]{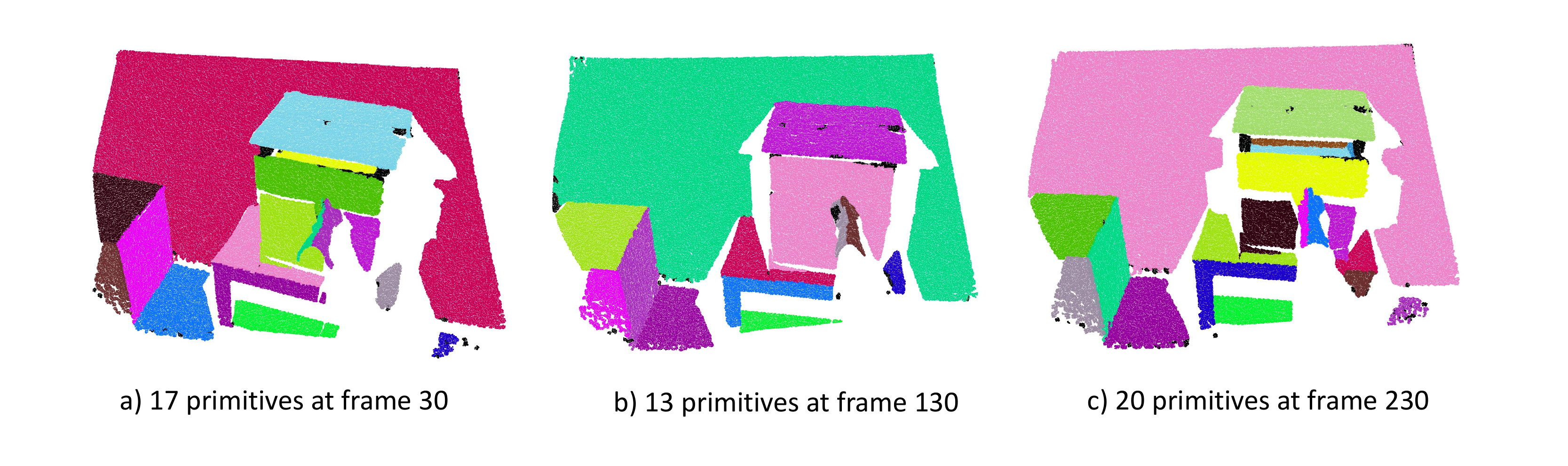}
    \caption{An illustration of primitive fitting in a HOI4D~\cite{liu2022hoi4d} video. Despite changing view angles and challenging interaction, the primitive fitting remains consistent across time.
    }
    \label{fig:stableFitting}
\end{figure}

To develop an efficient and effective backbone for long-term 4D understanding, we draw inspirations from the geometry processing community that primitive planes as some mid-level geometric representations are both compact and stable across time, see Figure~\ref{fig:stableFitting}.
Using primitive planes to model the long-term context not only eases the need to directly deal with the huge number of raw points in a 4D sequence but also facilitates long-term feature association. Furthermore, since primitive planes group points with coherent geometric features, it builds a natural geometry hierarchy (scene-primitive-point) which could be used as a strong inductive bias for powerful yet hard-to-optimize transformer-style architectures. We follow this thought and develop our Point Primitive Transformer. 

\begin{figure}[H]
    \centering
    \includegraphics[width=9cm]{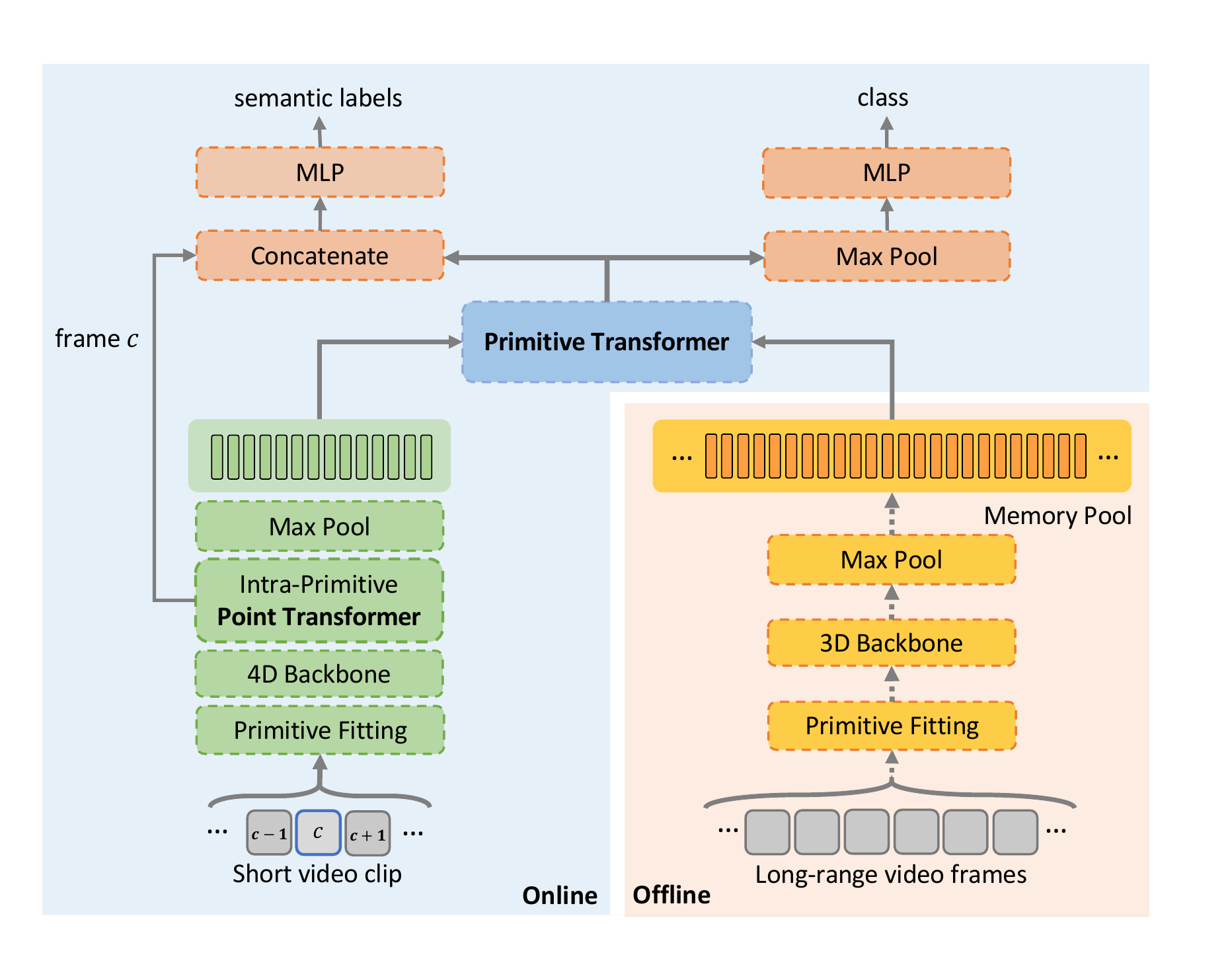}
    \caption{Pipeline. The backbone consists of two branches: online network and offline pre-computation. \textbf{1.Online Branch. }The input to network is a short video clip. After primitive fitting, points are tagged with primitive label, then 4D Backbone is applied and generates per-point features. In the  intra-primitive point transformer, points features are enhanced by adapatively adding information from other points inside primitive. Then generate primitive-level representations by maxpooling. In the primitive transformer, clip primitive embeddings(green) perform self-attention with long-term embeddings(yellow) in the memory pool. For semantic segmentation, primitive features are concatenated to corresponding point features then classified into semantic labels. For action recognition, primitive features are merged by maxpooling to a global feature then classified into actions. \textbf{2.Offline Branch. } This branch essentially computes primitive level representations of the long-range videos and maintains a memory pool in an offline manner. After primitive fitting, points are fed to a pre-trained 3D backbone. Then maxpool is applied to every primitive region generating primitive-level embeddings in the memory pool.}
    \label{fig:pipeline}
\end{figure}

Point Primitive Transformer(PPTr) is a two-level hierarchical transformer built upon the geometry hierarchy induced by primitive planes as shown in Figure~\ref{fig:pipeline}. On the lower level, short-term spatial-temporal features are extracted through an intra-primitive point transformer. The intra-primitive point transformer restricts the communication of points within each primitive plane. This design shares a similar flavor with GLOM~\cite{glom} encouraging aligned features to talk. Also due to the local spatial support, it is more friendly to optimization compared with a global transformer. On the upper level, long-term spatial-temporal features are extracted through a primitive transformer. This is done by jointly analyzing short-term features from the lower level and a memory pool storing pre-computed primitive features from a long video. Pre-computed primitive features allow aggregating long-term spatial-temporal context efficiently and effectively. PPTr is very flexible for both point-wise and sequence-wise inference by simply changing the task head.
In the rest of this section, we will elaborate on the design of PPTr in detail. We start with how we fit primitive planes and how we pre-compute primitive features in Section~\ref{sec:primitivefitting}. Then in Section~\ref{sec:intra-primitive} and Section~\ref{sec:inter-primitive}, we explain how we extract short-term and long-term spatial-temporal features respectively. 


\label{sec:hierarchicaltransformer}
\subsection{Primitive Fitting and Feature Pre-Computation}
\label{sec:primitivefitting}
We represent a point cloud sequence as $\Psi=\{(P_t, V_t)|t=1,\dots,  L\}$, where $P_t$ is the point cloud of frame $t$ optionally accompanied with normals $V_t$.
In this phase, we detect planes for each frame $(P_t, V_t)$ and output
primitive label $\Xi_t\in \mathbb{R}^{N\times3}$ and primitive parameters $\Theta_t\in \mathbb{R}^{M\times4}$, where $N$ is the number of points and $M$ is the number of primitives.
We adopt two primitive fitting methods in our study for different datasets: region grow~\cite{regiongrow1} and RANSAC~\cite{ransac}.

We leverage region grow for indoor and outdoor scene segmentation. Region grow detects planes based on normal estimation. If not provided with normal $V_t$, we calculate the normal direction at each point beforehand by linear least squares fitting of a plane over its nearest k neighbors.
Compared with region grow, RANSAC does not require normal estimation and is more suitable for low-resolution point clouds such as those for action recognition in MSR-Action3D~\cite{msr}.

After primitive fitting, we pre-compute the primitive features for efficient long-term context aggregation and form a memory pool $F_\text{mem}$ as shown in Figure~\ref{fig:pipeline}. Specifically, we pre-train a 3D point feature learner~\cite{p4transformer} to solve the task of interest just from every  single frame $(P_t,V_t)$. This allows us to extract per-point features $F_t\in\mathbb{R}^{C\times N}$ where $C$ denotes the feature dimension. To extract primitive level representations, point-wise max pooling is adopted for each primitive plane. The final memory pool $F_{\text{mem}}$ has a shape of $\mathbb{R}^{C\times M\times L}$.

\subsection{Short-Term Spatial-Temporal Feature Extraction}
This branch mainly consists of a 4D backbone and an intra-primitive point transformer.
The per-point features of each 4D sequence are first extracted using the 4D backbone. Following that, an intra-primitive point transformer is used to extract low-level features. Point features can provide the most fine-grained information, enabling us to better perform dense prediction tasks. The intra-primitive point transformer can not only align point features of similar geometry but also save computational overhead and reduce the optimization difficulty of the transformer.

\noindent\textbf{4D Backbone.}
Our 4D backbone is built using a UNet structure. Following the state-of-the-art P4Transformer~\cite{p4transformer}, the encoder/decoder is made up of four 4D convolution/decovolution layers. Given clip $\Psi$, the convolution layer can be described as:\\
\begin{equation}
\begin{aligned}
\boldsymbol{f}_{t}^{\prime(x, y, z)}
&=\sum_{\delta_{t}=-r_{t}}^{r_{t}} \sum_{\left\|\left(\delta_{x}, \delta_{y}, \delta_{z}\right)\right\| \leq r_{s}} 
(\boldsymbol{W_d}\cdot\left(\delta_{x}, \delta_{y}, \delta_{z}, \delta_{t}\right)^T) \odot(\boldsymbol{W_f} \cdot \boldsymbol{f}_{t+\delta_{t}}^{\left(x+\delta_{x}, y+\delta_{y}, z+\delta_{z}\right)})
\end{aligned}
\end{equation}
where $(x, y, z)\in P_t$ and $(\delta_x, \delta_y,\delta_z,\delta_t)$ is spatial-temporal offset of kernel and $\cdot$ is matrix multiplication. $f_t^{(x, y, z)} \in \mathbb{R}^{C\times1}$ is the feature of point at $(x, y, z, t)$, and the temporal aggregation $\sum$ is implemented with sum-pooling and the spatial $\sum$ is max-pooling. $r_s$ and $r_t$ represent temporal and spatial radius. $\boldsymbol{W_d}\cdot\left(\delta_{x}, \delta_{y}, \delta_{z}, \delta_{t}\right)^T$ generates offset weights where $\boldsymbol{W_d} \in \mathbb{R}^{C'\times 4}$ transforms 4D displacements from $\mathbb{R}^{4\times 1}$ to $\mathbb{R}^{C'\times 1}$, and $\boldsymbol{W_f} \in \mathbb{R}^{C'\times C}$ is a projection matrix. $\odot$ is summation. 

\noindent\textbf{Intra-Primitive Point Transformer.}
\label{sec:intra-primitive}
In this stage, the lower-level feature is extracted by enhancing per-point features obtained from the 4D backbone in a geometry-aware way. point features are clustered in groups according to their primitive labels given in the primitive fitting phase. Compared with simply grouping by k-NN search in euclidean space~\cite{pct}, primitive-based partition has a more underlying geometric meaning such as normal consistency. As point clouds are sets embedded in a metric space, self-attention is a natural way to build connections among them. By optimizing point embeddings in a geometry-aware manner, our intra-primitive transformer takes advantage of local aggregation rather than global information exchange. It is more friendly to optimization than a global transformer because points within the primitive plane cannot communicate with points outside. After this step, points with similar geometric features are easier to align together, which facilitates subsequent higher-level feature extraction. 
\begin{figure}[htbp]
    \centering
    \includegraphics[width=\linewidth]{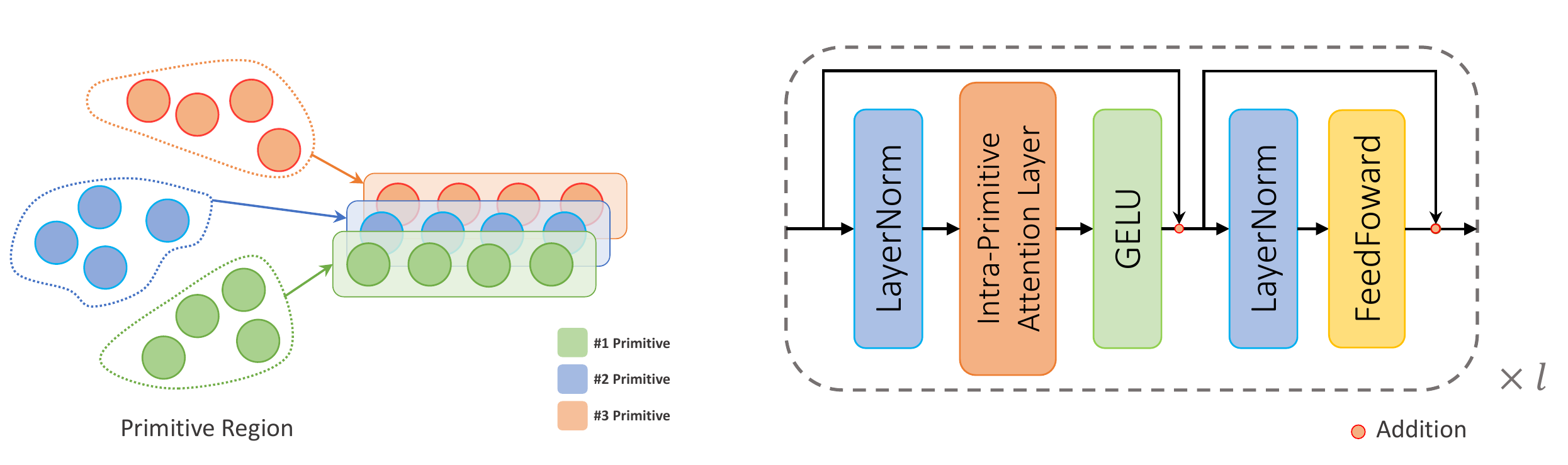}
    \caption{\textbf{Left: primitive-based region partition.} Points are divided into primitive regions according to primitive labels. Intra-primitive performs self-attention in one primitive region. \textbf{Right: Intra-primitive transformer block.} Consisting of intra-primitive attention layer, pre-LayerNorm~\cite{layernorm}, GELU~\cite{gelu} and residual connection~\cite{resnet}.}
    \label{fig:intra}
\end{figure}

Specifically, in the layer $i$, the enhanced point feature $F_{\text{out}}^i$ of primitive region $i$ with input embedding set  $F_{\text{in}}^{i}$ is computed formally as~\cite{attention}:
\begin{equation}
\begin{array}{cc}
     Q= W_q\cdot F_{in}^{i},  K = W_k\cdot F_{in}^{i},  V=W_v\cdot F_{in}^{i} \\
       F_{\text{out}}^{i}=SA( Q,  K,  V) = softmax\left(\frac{ Q^T K}{\sqrt{C^k}}\right)V  \\
\end{array}
\end{equation}
where $F_{\text{in}}^i \in \mathbb{R}^{C^i \times N' \times L M}$, $C^i,N',L,M$ represents input dimension, point number per primitive, clip length and primitive number  respectively. $W_q, W_k \in \mathbb{R}^{C^k\times C^i}$, $W_v \in \mathbb{R}^{C^v\times C^i}$ where $C^k$ is the key dimension and $C^v$ is value dimension. $Q, K, V$ are queries, keys and values generated from $F_{in}^{i}$. Attention weights  $softmax\left(\frac{ Q^T K}{\sqrt{C^k}}\right)$ is calculated in the primitive region. The output $F_{\text{out}}^i \in \mathbb{R}^{C^v \times N' \times L M} $ is computed as a weighted sum of the values V. 
As shown in Figure~\ref{fig:intra}, we build intra-primitive transformer block with layernorm~\cite{layernorm}, GELU activation~\cite{gelu}, one attentive layer and a following feedforward layer~\cite{attention}. Feedforward is implemented with a two-layer MLP(MultiLayer Perception). 

\subsection{Long-Term Spatial-Temporal Feature Extraction}
\label{sec:inter-primitive}
After the short-term spatial-temporal feature extraction, primitive transformers are used to jointly analyze short-term features from the lower level and a memory pool containing pre-computed primitive features. This branch can not only reduce the computational cost, but also achieve long-term spatio-temporal information integration. 

\noindent\textbf{Primitive Transformer.}
As demonstrated in Fig~\ref{fig:pipeline}, two branches merge here. The output of $l$ layer intra-primitive transformer $F_{out}^l \in \mathbb{R}^{C^l\times N'\times L M}$ is then aggregated by max-pooling operator $MAX\{\cdot \}$ to obtain primitive level feature $F_{out} \in \mathbb{R}^{C^l\times L M}$, where $C^l$ is feature channels, $L$ is the clip length and $M$ is the primitive number. Pre-computed primitive features from memory pool $F_{\text{mem}}$  are used to expand the spatio-temporal receptive field of the primitive transformer. 
Formally, the input of primitive transformer is $F_{\text{in}}^{\text{primitive}} = [F_{\text{clip}} ||  F_{\text{mem}}] \in \mathbb{R}^{C^l\times (L'+L)M}$ which concatenates short-term primitive features $F_{\text{clip}}$ and  primitive features from memory pool $F_{\text{mem}}$. Note that in the primitive attention layer, spatial-temporal attentive aggregation is performed in $(L'+L)\times M$ primitive regions simultaneously. Identical to intra-primitive shown in Fig~\ref{fig:intra}(Right), primitive transformer block is also composed of pre-LayerNorm, primitive attention layer, GELU, feedforward layer and residual connection. 
For semantic segmentation, we concatenate per-point features, intra-primitive point features, and primitive features to obtain point-wise features, and fuse them by a three-layer MLP. For the action recognition task, we use the primitive feature to obtain classification predictions through max-pooling and MLP.



\section{Experiments}
\subsection{4D Semantic Segmentation}
\textbf{Setup.} Temporal information can help understand the dynamic objects in the scene, and improve segmentation accuracy and robustness to noise. Due to memory constraints, existing methods only process point cloud videos with a length of 3. Our method can consider a longer temporal range and achieve a more efficient integration of spatio-temporal information. In this task, we fit the scene point cloud into 200 primitives. 
We use mean IoU(mIoU) \%  as the evaluation metric.

\subsubsection{4D Semantic Segmentation on Synthia 4D dataset.}
\textbf{Setup.} Synthia 4D~\cite{synthia} is a synthetic dataset for outdoor autonomous driving. It creates 3D videos with the Synthia dataset, which consists of six videos of driving scenarios in which objects and cameras are moving. We use the same training/validation/test split as previous work, with 19,888/815/1,886 frames, respectively.

\newcommand{\tabincell}[2]{\begin{tabular}{@{}#1@{}}#2\end{tabular}} 
\begin{table}
\setlength{\tabcolsep}{0.2mm}
\tiny{
\begin{center}
\caption{Evaluation for semantic segmentation on Synthia 4D dataset~\cite{synthia}}
\label{table:synthia4d}
\begin{tabular}{l|c|cccccccccccc|c}
\hline\noalign{\smallskip}
Method & Frames & Bldn& Road& Sdwlk & Fence& Vegittn & Pole & Car & T.Sign & Pedstrn & Bicycl & Lane & T.Light& mIoU\\
\noalign{\smallskip}
\hline
\noalign{\smallskip}
3D MinkNet14~\cite{minkowski}&1 & 89.39& 97.68&69.43 &86.52 &98.11 &97.26 &93.50 &79.45 &92.27 & 0.00 & 44.61 & 66.69 & 76.24\\
4D MinkNet14~\cite{minkowski}& 3 & 90.13& 98.26& 73.47& 87.19& 99.10& 97.50& 94.01& 79.04& \textbf{92.62}& 0.00 & 50.01 & 68.14 & 77.24\\
\hline
\noalign{\smallskip}
PointNet++~\cite{pointnet2} & 1& 96.88 &  97.72 & 86.20 & 92.75 & 97.12 & 97.09 & 90.85 & 66.87 & 78.64 & 0.00 & 72.93 & 75.17 & 79.35 \\
MeteorNet-m~\cite{meteornet} & 2 & \textbf{98.22} & 97.79 & 90.98 & 93.18 & 98.31 & 97.45 & 94.30 & 76.35  & 81.05 & 0.00 & 74.09 & 75.92& 81.47 \\
MeteorNet-l~\cite{meteornet} & 3 & 98.10 & 97.72  & 88.65 & 94.00 & 97.98 & 97.65 & 93.83 & \textbf{84.07}  & 80.90  &0.00 & 71.14 & 77.60 & 81.80\\
\noalign{\smallskip}
\hline
\noalign{\smallskip}
P4Transformer~\cite{p4transformer} & 1 & 96.76 & 98.23 & 92.11 & 95.23 & 98.62 & 97.77 & 95.46 & 80.75 & 85.48 & 0.00 & 74.28 & 74.22 & 82.41\\
P4Transformer~\cite{p4transformer} & 3 & 96.73 & 98.35 & 94.03 & 95.23 & 98.28 & 98.01 & 95.60 & 81.54 & 85.18 & 0.00 & 75.95 & \textbf{79.07} & 83.16 \\
\noalign{\smallskip}
\hline
\noalign{\smallskip}
\tiny{PPTr(ours)} & 1 &97.14 &98.42 &94.12 & 97.00&99.59 & 97.86 & 98.54 & 79.68 & 89.20 & 0.00 & 77.26 & 77.42 & 83.85\\
\tiny{PPTr(ours)} & 30 &98.01 &\textbf{98.63} &\textbf{95.26} &\textbf{97.03} & \textbf{99.70}&\textbf{97.95} &\textbf{98.76} & 81.99&91.20&0.00&\textbf{78.29}&77.09&\textbf{84.49}\\
\noalign{\smallskip}
\hline
\end{tabular}
\end{center}
}
\end{table}

\textbf{Result.} Table~\ref{table:synthia4d} shows our method outperforms the state-of-the-art methods. Our PPTr with 1 frame can achieve $0.69\%$ improvement over the P4Transformer with 3 frames, which demonstrates the effectiveness of the hierarchical structure. When using the memory pool to integrate temporal information from 30 frames, we can achieve $1.33\%$ improvement over previous state-of-the-art methods. It is worth mentioning that our method is the first to integrate point clouds of 30 frames, which is 10 times that of previous methods. And we also demonstrate that longer point cloud sequences are valuable for 4D semantic segmentation.

\subsubsection{4D Semantic Segmentation on HOI4D.}
\textbf{Setup.} In order to further verify the effectiveness of our method, we select the HOI4D dataset for experiments, which is a large-scale 4D egocentric dataset to catalyze the research of category-level human-object interaction. It provides frame-wise annotations for 4D point cloud semantic segmentation. Since the dataset has not been released yet, we sent an email to the author team to request 1000 sequences, which includes 30k frames of the point cloud. The train/test split is the same as HOI4D.


\begin{table}
\setlength{\tabcolsep}{0.15mm}
\scriptsize{
\begin{center}
\caption{Evaluation for semantic segmentation on HOI4D dataset~\cite{liu2022hoi4d}}
\label{table:hoi4d}
\begin{tabular}{l|c|ccccccccc|c}
\hline\noalign{\smallskip}
Method & Frames  & Table &Ground  & Metope &Locker & Pliers  &Laptop &\tabincell{c}{Safe\\Deposit} &Pillow &\tabincell{c}{Hand \\and Arm}& mIoU\\
\noalign{\smallskip}
\hline
\noalign{\smallskip}
\tiny{PSTNet~\cite{pstnet}} & 3&	57.45 &	63.38 &	83.80 &	44.69 &	13.71 &	35.03 &	51.55 &	76.30 &	40.39 &51.81 \\
\tiny{P4Transformer~\cite{p4transformer}} & 1 & 60.84 &  71.98   &	86.69 &	53.89 &	34.00 &	65.89 &	55.87 &	52.19 &	55.10 & 59.61 \\
\tiny{P4Transformer~\cite{p4transformer}} & 3 & 63.58 &	66.60 &	87.17 &	58.39 &	32.29 &	72.03 &	65.87 &	57.41 &	54.36 & 61.97 \\
\noalign{\smallskip}
\hline
\noalign{\smallskip}
\tiny{PPTr(ours)} & 1&	67.49 &	74.92 &	87.92 &	\textbf{62.12} &	40.06 &	69.00 &	71.39 & 77.18 &	62.50 & 68.07 	 \\
\tiny{PPTr(ours)} & 3& 66.78 &	72.76 &88.21	 &	60.83 &	\textbf{41.22} &	72.04 &	73.10 &	80.64 &	61.27 & 68.54\\
\tiny{PPTr(ours)} & 30& \textbf{67.76} &	\textbf{79.55} &\textbf{90.67}	 &	59.43 &	39.43 &	\textbf{72.67} &	\textbf{73.29} &	\textbf{84.13} & \textbf{64.26} & \textbf{70.13}\\
\noalign{\smallskip}
\hline
\end{tabular}
\end{center}
}
\end{table}


\textbf{Result.} As shown in Table~\ref{table:hoi4d}, our method outperforms previous methods on this more challenging dataset. Compared with P4transformer, the mIoU goes up from $59.61\%$ to $68.07\%$ and $61.97\%$ to $68.54\%$ in the case of single frame and 3 frames respectively, demonstrating the effectiveness of the hierarchical design again. Due to the limitation of computational resources, P4Transformer can use up to 3 frames, but our method can integrate 30 frames of spatio-temporal information. The improvement from $61.97\%$ to $70.13 \%$ further confirms that with our proposed primitive memory pool, we can better leverage the long-term temporal information to boost the 4D segmentation performance. 

\subsection{3D Action Recognition on MAR-Action3D}
\textbf{Setup.} To demonstrate the effect of PPTr, we first conduct experiments on the 3D Action Recognition task. Followed by P4Transformer, we use the MAR-Action3D dataset which consists of 567 human body point cloud videos, including 20 action categories. Our test/train split follows previous work. Each frame is sampled with 2,048 points. As inputs, point cloud videos are split into multiple clips. Video-level labels are used as clip-level labels during training. In order to estimate the video-level probability, we take the mean of all clip-level probability predictions. We fit the human body point cloud into 4 primitives. 
Due to the small scale of human point cloud videos, we can load the entire point cloud videos at one time, so we can avoid maintaining the long-term memory pool in this case. We use the video classification accuracy as the evaluation metric. We compare our method with the latest 4D backbone for point cloud video including MeteorNet, PSTNet and P4Transformer.

\begin{table}
\setlength{\tabcolsep}{1mm}
\scriptsize{
\begin{center}
\caption{Evaluation for action recognition on MSR-Action3D dataset~\cite{msr}}
\label{table:msraction3d}
\begin{tabular}{l|c|c|c}
\hline\noalign{\smallskip}
Method & Input & Frames & Accuracy \\
\noalign{\smallskip}
\hline
\noalign{\smallskip}
PointNet++~\cite{pointnet2}& point& 1 & 61.61 \\
\noalign{\smallskip}
\hline
\noalign{\smallskip}
& point& 4 & 78.11\\
& point& 8 & 81.14\\
MeteorNet~\cite{meteornet}& point& 12 & 86.53\\
& point& 16 & 88.21\\
& point& 24 & 88.50\\
\noalign{\smallskip}
\hline
\noalign{\smallskip}
& point& 4 & 81.14\\
& point& 8 & 83.50\\
PSTNet~\cite{pstnet}& point& 12 & 87.88\\
& point& 16 & 89.90\\
& point& 24 & 91.20\\
\noalign{\smallskip}
\hline
\noalign{\smallskip}
& point& 4 & 80.11\\
& point& 8 & 83.17\\
P4Transformer~\cite{p4transformer}& point& 12 & 87.54\\
& point& 16 & 89.56\\
& point& 24 & 90.94\\
\noalign{\smallskip}
\hline
\noalign{\smallskip}

& point& 4 & 80.97 \\
& point& 8 & 84.02\\
PPTr(ours)& point&12  & 89.89\\
& point&  16& 90.31\\
& point&  24& \textbf{92.33}\\
\noalign{\smallskip}
\hline

\end{tabular}
\end{center}
}
\end{table}

\textbf{Result.} As reported in Table~\ref{table:msraction3d}, when the number of point cloud frames increases, the classification accuracy can be gradually improved. Our method outperforms all the state-of-the-art methods, demonstrating that our methods can better integrate spacial-temporal information.

\subsection{Ablation Study and Discussion}
In this section, we first provide an ablation study to verify each component. Then, we provide more analysis to provide an in-depth understanding of our framework.

\subsubsection{Efficacy of intra/inter-primitive Transformer.}
We run ablation studies with and without intra/inter-primitive Transformer to quantify its efficacy. We find that PPTr without intra/inter-primitive Transformer results in a $16.73$/$1.39$ accuracy drop on the MSR-Action3D action recognition task. This shows that the intra-primitive transformer is essential in this task. It not only simplifies the optimization difficulty but also aligns similar points, providing good features for the subsequent use of the inter-primitive transformer. Inter-primitive Transformer integrates spatio-temporal information from the entire video, using the complementary information of each frame to further improve classification accuracy.

\subsubsection{Robustness to primitive-fitting hyper-parameters.}

The performance impacts of different numbers of primitives are provided since primitives are crucial in the framework. On MSR-Action3D, we can achieve $91.5$/$91.89$ accuracy with 2/8 primitives, resulting in a marginal drop. On Synthia 4D, the segmentation mIoUs are $82.98$, $84.41$, $84.49$, $84.28$, and $83.56$ with a primitive number of 10, 100, 200, 400, and 2000 respectively. Notice when the primitive number varies in a reasonable range from 100 to 400, the segmentation mIoUs vary by no more than 0.21. When the primitive number is 10, the region division is too coarse for fine-grained segmentation.When the primitive number is 2000, the benefit of the spatial hierarchy gets weakened a lot. The network degenerates to a point transformer when further increasing the primitive number to the point number. This shows that different numbers of primitives have a small effect on the results, and all have consistent improvements.

\subsubsection{Efficacy of primitive representation.}


Our hierarchical transformer is generic and can be easily applied to mid-level representations other than primitive planes.
To confirm the efficacy of primitive planes, we additionally compare primitive planes with two types of mid-level representations, BPSS~\cite{lin2018toward} supervoxels and k-means clusters.
Results in the table below show that using BPSS supervoxels outperforms P4Transformer but is not as good as using primitive planes while k-means clusters fail to serve as a beneficial mid-level representation on Synthia4D.
\begin{table}[h]
\begin{center}
\caption{Comparisons between different representations}
\label{table:supervoxel}
\begin{tabular}{ccc}
\hline\noalign{}
 Method & Synthia 4D~\cite{synthia} & MSRAction~\cite{msr} \\
\hline\noalign{}
P4Transformer & 83.16 & 90.94\\\noalign{}
K-means &80.70& 91.76\\\noalign{}
BPSS~\cite{lin2018toward} &83.43&91.98\\\noalign{}
Ours &\textbf{84.49} & \textbf{92.33}\\\noalign{}

\hline
\end{tabular}
\end{center}

\end{table}
\subsubsection{Offline branch and Online branch.} The online branch produces fine primitive features with heavy computation while the offline branch produces coarser features efficiently as a surrogate of the online branch so that the network can process long clips with limited computing resources.
For the action recognition task where data clips can already be largely fit into the GPU memory, using an online branch only with fine primitive features is preferred. In this case, just using an offline branch or combining the offline and online branches results in marginal performance degradation with accuracy of 92.13 and 92.27 respectively. For the 4D segmentation task, using our online branch independently, the memory could only afford 3 frames and the resulting segmentation mIoU(\%) is 84.05. This number goes to 84.49 when assisted by the offline branch covering 30 frames, confirming the value of the offline branch. 


\section{Conclusions}
This paper proposes a 4D backbone for long-term point cloud video understanding. The key idea is to leverage the primitive plane to capture the long-term spatial-temporal context in 4D point cloud videos. Results of experiments showing ablations and state-of-the-art performance on a wide range of 4D tasks including MSR-Action3D action recognition task, 4D semantic segmentation on sythia4D and on HOI4D. This result is very encouraging and suggests future work to explore more possible backbone designs for 4D point cloud understanding. 
%
%
\bibliographystyle{splncs04}
\bibliography{egbib}
\end{document}